\pdfoutput=1

\documentclass[10pt,conference]{IEEEtran}
\usepackage{times}

\usepackage{graphicx}
\usepackage{epstopdf}
\usepackage{multirow, makecell}
\usepackage{balance}
\usepackage{enumitem}
\usepackage{url}

\usepackage{amsmath}

\newcommand\blfootnote[1]{%
  \begingroup
  \renewcommand\thefootnote{}\footnote{#1}%
  \addtocounter{footnote}{-1}%
  \endgroup
}

\renewcommand{\footnoterule}{%
  \kern -3pt
  \hrule width \columnwidth height 1pt
  \kern 2pt
}

\usepackage{color}
\begin{document}

\pagenumbering{gobble}


\title{\textbf{A Reproducible Comparison of RSSI Fingerprinting Localization Methods Using LoRaWAN}\\}

\author{\IEEEauthorblockN{~\large Grigorios G. Anagnostopoulos, Alexandros Kalousis}
\IEEEauthorblockA{Geneva School of Business Administration, HES-SO
Geneva, Switzerland\\
Email:  $\left\{ grigorios.anagnostopoulos, alexandros.kalousis \right\}$@hesge.ch}
}

\maketitle

\begin{abstract}
The use of fingerprinting localization techniques in outdoor IoT settings has started to gain popularity over the recent years. Communication signals of Low Power Wide Area Networks (LPWAN), such as LoRaWAN, are used to estimate the location of low power mobile devices. 
In this study, a publicly available dataset of LoRaWAN RSSI measurements is utilized to compare different machine learning methods and their accuracy in producing location estimates.  The tested methods are:  the k Nearest Neighbours method, the Extra Trees method  and a neural network approach using a Multilayer Perceptron. To facilitate the reproducibility of tests and the comparability of results, the code and the train/validation/test split of the dataset used in this study have become available. The neural network approach was the method with the highest accuracy, achieving a mean error of 358 meters and a median error of 204 meters.
\end{abstract}

\renewcommand\IEEEkeywordsname{Keywords}

\IEEEpeerreviewmaketitle

\blfootnote{\\
Code: To be uploaded in zenodo.org\\
Data: To be uploaded in zenodo.org\\
}

\section{Introduction} \label{sec:Introduction}

The proliferation of the usage of Internet-of-Things (IoT) technologies and Low Power Wide Area Networks (LPWAN), such as  LoRaWAN or Sigfox, over the last decade has created a new landscape in the field of outdoor localization.  Low power devices of  LPWANs cannot afford the battery consumption of a chip-set of a Global Navigation Satellite System (GNSS), such as the GPS. Therefore, an alternative approach is needed in order to localize these low power devices.  

The devices communicate with fixed basestations deployed in urban and rural areas through RF messages. These messages contain the information that is needed to be collected, which is application specific. Moreover, there is additional information in these signal transmissions and receptions, which can be utilized for localization. Measurements such as the Received Signal Strength Indicator (RSSI), the Time Difference of Arrival (TDoA) or the Logarithmic Signal over Noise Ratio (LSNR) can be used for producing location estimates.
The signals can be used either by a ranging method (such as multilateration) or by a fingerprinting method. Fingerprinting methods have the downside that they require the tedious step of collecting a dataset of fingerprints. This is compensated, however, by the fact that they generally offer better accuracy that the ranging methods, since they capture the particularities of the environment.

In the current study, a public dataset of LoRaWAN RSSI fingerprints is utilized to perform a reproducible comparison of different machine leaning methods. The common fingerprinting technique used, namely k-nearest neighbours, is compared with two other methods: the Extra Tress method, and a Neural Network. For each method, a non-exhaustive examination and tuning of their most important hyperparameters is performed.
 
The rest of this paper is organized as follows. In Section~\ref{sec:Related}, the related work is discussed. Section~\ref{sec:Dataset} presents in detail the dataset used in this work. In Section~\ref{sec:Preprocessing}, the preprocessing steps studied in this work are presented. The concise description of the experimental setup in Section~\ref{sec:Setup} is followed by an extensive presentation and discussion of the results in  Section~\ref{sec:Results}. Finally, Section~\ref{sec:Conclusions} summarizes this work with a short discussion of the drawn conclusions.

\section{Related Work} \label{sec:Related}

Fingerprinting has been a broadly studied method of indoor positioning~\cite{WiFi_Fingerprint_Overview}.
More particularly, RSSI has been the main type of signal that is used ~\cite{WiFi_Fingerprint_Overview}. It has been only a few years since the transfering of fingerprinting techniques in the outdoor world, and in particular in LPWAN settings. In a recent study, Aenrouts et al.~\cite{Sigfox_Dataset}  have made three fingerprinting datasets of Low Power Wide Area Networks publicly available. One of these datasets contains LoRaWAN RSSI measurements collected in the urban area of the city of Antwerp, in Belgium. The motivation for making the datasets publicly available was to provide the global research community  with  a benchmark  tool  to  evaluate  fingerprinting algorithms  for LPWAN standards.  In that work, the utilization of the presented LoRaWAN dataset by a k Nearest Neighbours fingerprinting method was exemplified, achieving a mean localization error of 398 meters. To the best of our knowledge, there is no follow up study so far, which utilizes this dataset.

Plets et al.~\cite{Plets}, have evaluated experimentally RRS and TDoA ranging positioning methods using a LoRaWAN network, reporting median errors of 1250 and 200 meters for RRS and TDoA respectively. Other works~\cite{Choi},\cite{Gotthard}, have focused on rather specific settings over which they evaluate positioning methods. These works~\cite{Choi},\cite{Gotthard},  present experiments in  car parking settings, testing in confined areas, with a placement of basestations that was adapted to their use-case, presenting a low error which ranges at the scale of a few tens of meters.

General purpose fingerprinting methods in LPWAN settings have been presented and discussed in recent works~\cite{Janssen_Sigfox},~\cite{IPIN2019}. Janssen et al.~\cite{Janssen_Sigfox} have utilized a Sigfox dataset to apply a kNN algorithm, and selected among a variety of distance metrics and data representations the best performing ones, resulting with a mean positioning error of 340 meters. In addition, in our previous work~\cite{IPIN2019}, we have moved further in analysing the same Sigfox dataset, by tuning relevant parameters of the discussed preprocessing schemes, reducing the mean error to 298 meters. As was done in our previous work ~\cite{IPIN2019}, in order to facilitate the comparability of results, we share the train/validation/test sets used in the current work as well.

\section{The Dataset Used} \label{sec:Dataset}

\begin{figure}[!h]
\centering
\includegraphics[width=0.99\linewidth]{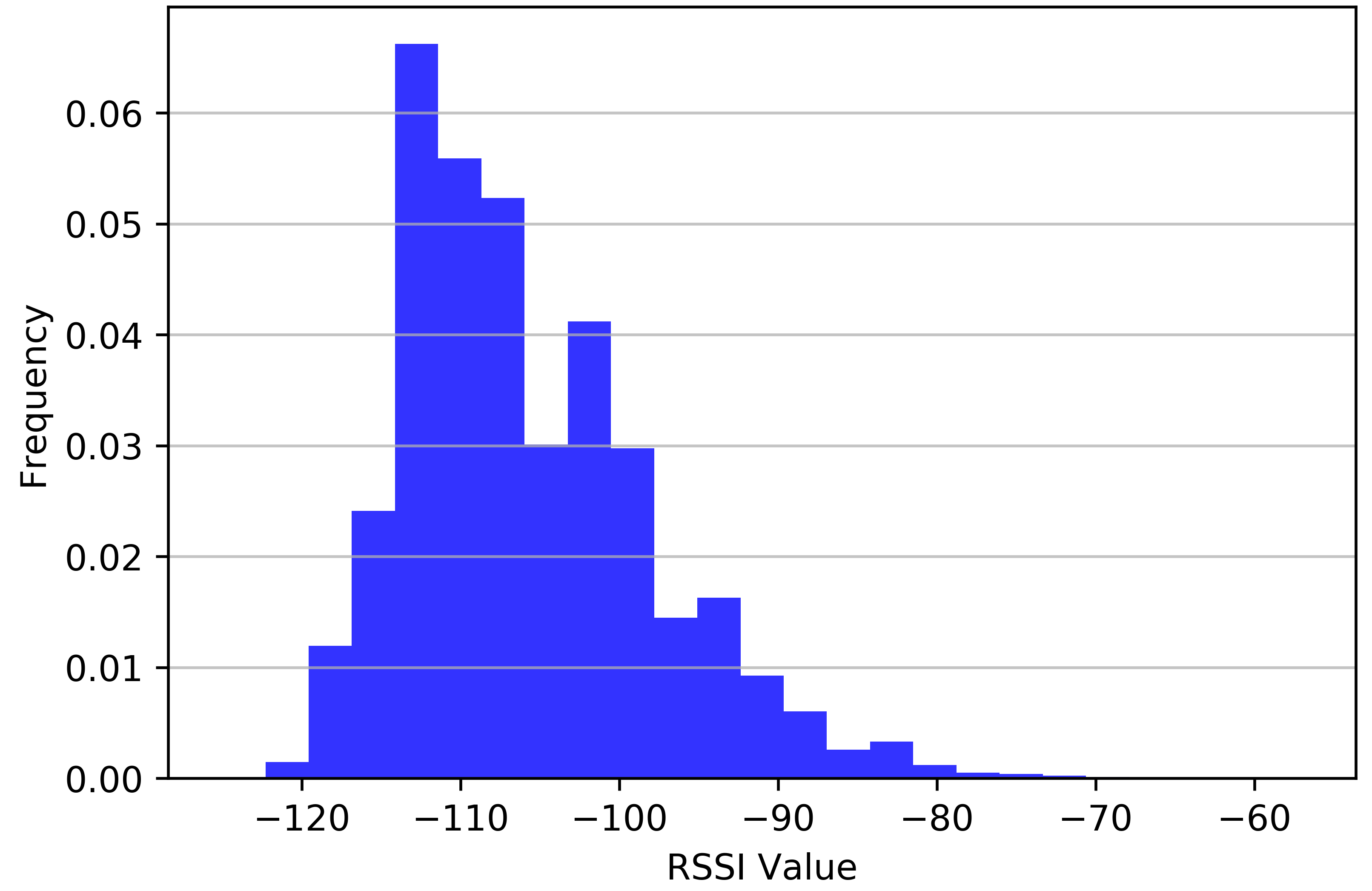}
\caption{The histogram of RSSI values of all signal receptions.
} \label{fig:histogram_rssi_values.png}
\end{figure}

The LoRaWAN dataset used in the current study has been published by Aenrouts et al.~\cite{Sigfox_Dataset}.
The fingerprints are collected in the urban area of Antwerp, in Belgium. A total number of 123528 messages are reported in the dataset. Each message contains the following information:
 the RSSI value of the transmitted signal as received by each of the 68 basestations, and the spatial ground truth of the signal's transmission location, as estimated by a GPS device, hold alongside the LoRaWAN devices. Additionally, the LoRa spreading factor, which is a parameter related to the signal's modulation and affects the range of the signal, is also reported. Lastly, the  Horizontal Dilution Of Precision (HDOP) of the GPS estimates is also reported. It is noteworthy that the ground truth is determined by a GPS estimate, which can be inaccurate. This is a known limitation of this setting. Nevertheless, the achievable error of RSSI localization with LoRaWAN is at least one order of magnitude greater that the one of GPS.
The histogram of the RSSI values of the dataset is presented in Figure~\ref{fig:histogram_rssi_values.png}, offering an overview of the RSSI values and their distribution. For all basestations that did not receive a signal's transmission, an out of range value $-200$ was artificially inserted by the creators of the dataset, which evidently is not included in the histogram.

\begin{figure}[!t]
\centering
\includegraphics[width=0.99\linewidth]{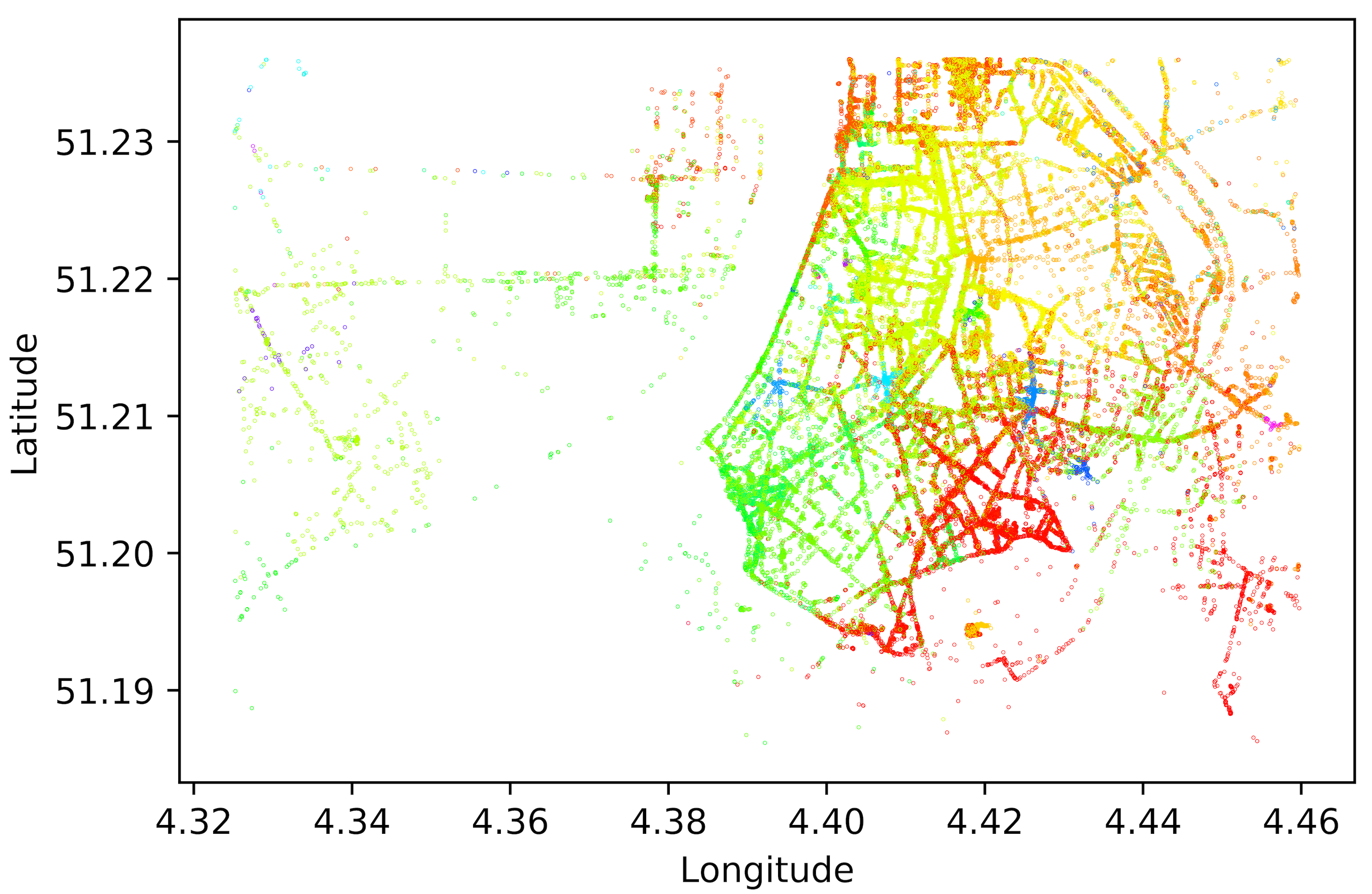}
\caption{The spatial distribution of the data points of the dataset.
} \label{fig:map.png}
\end{figure}

The dataset is quite sparse, as most signal transmissions were received by exactly three basestations, which is the minimum required number of gateways for having a unique solution when using a multilateration method. The details of the number of receiving basestations in all signal transmissions of the dataset are presented in Table~\ref{Table:gateways}. The accumulation of most points at the slot of exactly three receiving gateways is rather surprising. This could potentially be the result of some selection process or some practical limitation of the recording process of the signal receptions.  Overall, improving the accuracy of position estimates using machine learning with such a sparse dataset is a rather challenging task.

\begin{table}[h] \caption{Amount of Signal Receptions for Various a Number of Receiving Gateways}\label{Table:gateways}
\centering
\begin{tabular}{ |c|c| } 
 \hline
 \textbf{Number of  gateways $G$} & \textbf{Messages received by $G$ gateways} \\  \hline
 1 & 93  \\  \hline
 2 & 9424 \\  \hline
 3 & 113972  \\  \hline
 4 & 2  \\  \hline
 5 & 16  \\  \hline
 6 & 21  \\  
 \hline
\end{tabular}
\end{table}

The spatial distribution of the dataset is presented in Figure~\ref{fig:map.png}. Each color-defined cluster represents data points with the same basestation as the closest detected one (the one with the highest RSSI value).

For the purposes of our study, we have randomly split the dataset into a training, a validation and a test set containing 70\%, 15\%, and 15\% of the sample respectively (86469, 18529 and 18530 entries). 
The three subsets used in the current study are made available to the research community for future reference.

\section{Data Preprocessing} \label{sec:Preprocessing}  

In this work, four data prepossessing schemes are evaluated, which have been comonly used in previous works~\cite{TORRES_SOSPEDRA_2015},~\cite{Janssen_Sigfox},~\cite{IPIN2019}. The first two are: the \textit{positive} and the \textit{normalized} data representations. In the positive data representation, all out-of-range values are set to zero, and from all received RSSI values the minimum RSSI value found in the training set reduce by 1, is subtracted ($RSSI_i-(RSSI_{min}-1)$). With this representation, all out-of-range values are equal to zero and the receptions have positive values, from 1 and above. It is also possible to use a threshold $\tau$ as the minimum meaningful measured value. In the case of using the threshold $\tau$, all values bellow $\tau$ are set to zero, as all the out-of-range values as well. In the current study, the threshold $\tau$ is not used, thus all received RSSI values are shifted to a positive value range. The \textit{normalized} data representation is rescaling the \textit{positive} values into the [0,1] range.

The last two data representations, proposed by Torres-Sospedra et al.~\cite{TORRES_SOSPEDRA_2015}, are the \textit{exponential} and \textit{powed} data representations. The motivation for proposing these non-linear transformations is to go beyond the linear handling of the RSSI values, since the RSSI values correspond to a logarithmic scale. The \textit{exponential} and \textit{powed} representations are defined as follows:

\begin{equation}
        Exponential_i\\(x)= \frac{exp( \frac{Positive_i\\(x)}{\alpha})}{exp( \frac{-min}{\alpha})}
        \label{Eq:Exponential}
\end{equation}

\begin{equation}
       Powed_i\\(x)= \frac{(Positive_i\\(x))^\beta}{(-min)^\beta)}
        \label{Eq:Powed}
\end{equation}

In Equations~\ref{Eq:Exponential} and~\ref{Eq:Powed}, $min$ coresponds to the minimum RSSI in the training set, while $\alpha $ and $\beta$ are parameters to be defined. 
The proposed default values~\cite{TORRES_SOSPEDRA_2015} for the parameters $\alpha $ and $\beta $ are  $\alpha=24 $ and $\beta = e $, where $e$ is the mathematical constant.
Nevertheless, it is recommended to adjust these parameters to the distribution of the RSSI values of each different setting. In the case of this study, it is particularly recommended to do so, since the default parameters where proposed upon experimentation with Wi-Fi signals in an indoor setting, while in the current study an outdoor LoRaWAN setting is studied.

\section{Experimental Setup} \label{sec:Setup}

 For the experiments presented in this study, the free machine learning library for the Python language, scikit-learn, was used. Particularly, scikit-learn version 0.19.1 and Python 3.5.5 were used. Additionally, for the Neural Network tests, the Keras (2.2.4) deep learning open-source library was used. The Haversine formula has been used for measuring distances on the reported experiments. 

The common train/validation/test set methodology is used on the analyses presented in this work. These sets, as well as the code of the tests of this study can be accessed here (in the camera ready version they will be uploaded to zenodo.org, currently in https://www.dropbox.com/sh/6xs6qithusmb4s9/ AACN6fFc4cgUmNzPtyn7w7mUa?dl=0).
 For the kNN method, the RSSI values received by the 68 basestations present in the dataset are used as the input features. The two other methods (Extra Trees and MLP), apart from the RSSI values, also utilize as an additional feature the reported LoRa spreading factor. This distinction is made because, contrary to kNN, Trees and Neural Nets can naturally process non-commensurable features. 
For each of the three methods, a selection of appropriate values is made for the most important hyperparameters, so as to end up with a model that is able to generalize well. Moreover, the preprocessing schemes presented in Section~\ref{sec:Preprocessing} are studied. In the experiments of the first method, the kNN, a detailed comparison and a tuning of the parameters of these preprocessing schemes is carried out. The examination of these four preprocessing schemes has been made with all three machine learning methods. Contrary to previous works that study other datasets~\cite{IPIN2019},~\cite{Janssen_Sigfox},~\cite{TORRES_SOSPEDRA_2015}, the results here do not indicate big differences among these preprocessing schemes for the given dataset.

\section{Tests and Results} \label{sec:Results}

\subsection{The k-nearest neighbours method} \label{sec:knn}

In this section, the kNN method, which is the most common fingerprinting method used, is evaluated. Initially, the parameters $\alpha$ and $\beta$ of the two preprocessing methods discussed in Section~\ref{sec:Dataset} (in Equations~\ref{Eq:Exponential} and~\ref{Eq:Powed}) are tuned. Then, all distance metrics available in the scikit-learn toolkit are evaluated using the four prepossessing methods previously discussed. For each setting, the value of the hyperparameter  $k$ that minimizes the mean error on the validation set is selected.

\subsubsection{Tuning Preprocessing Parameters $\alpha$ and $\beta$} \label{sec:a_b}

\begin{figure}[!h]
\centering
\includegraphics[width=0.95\linewidth]{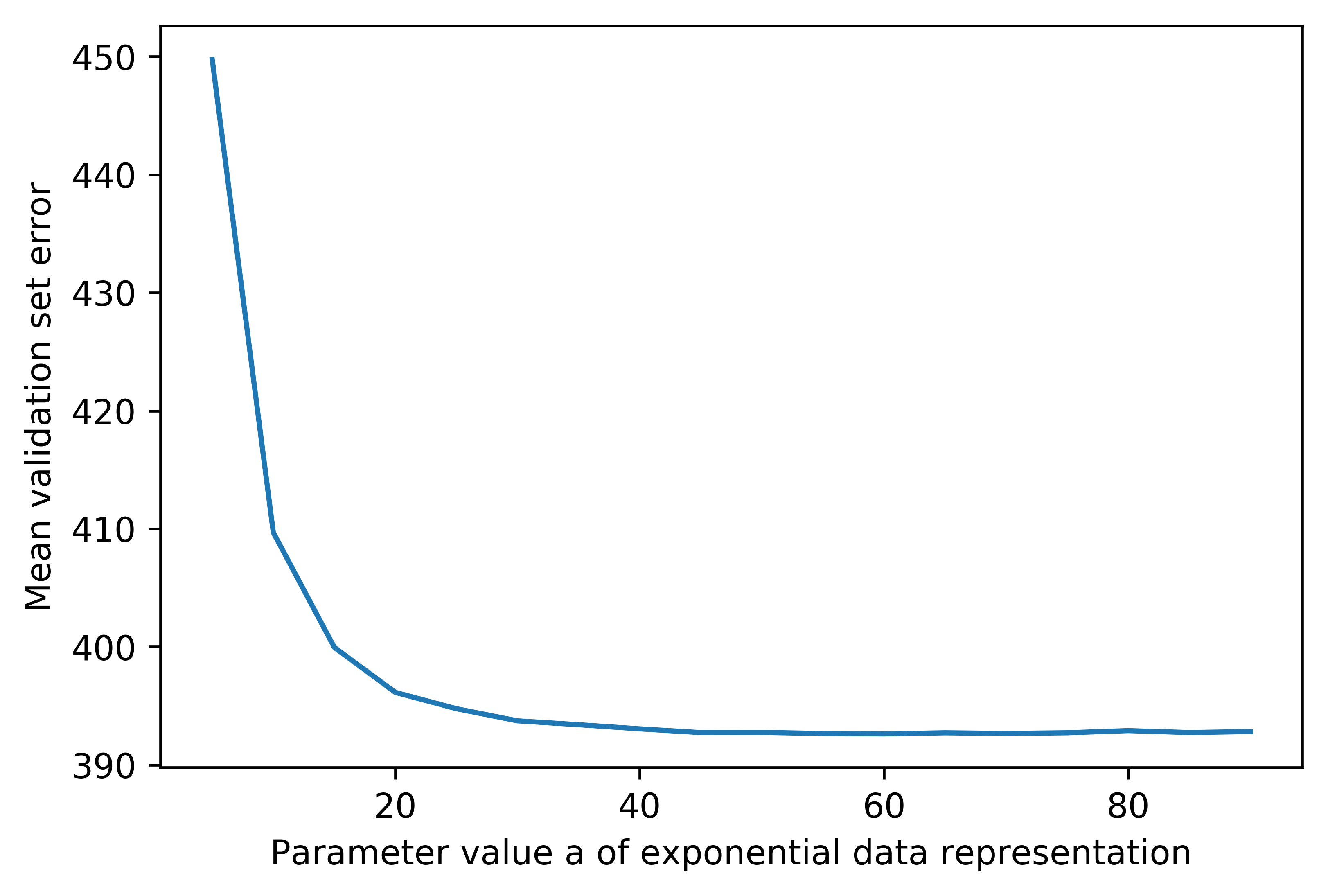}
\caption{The mean error on the validation set for different values of the parameter $\alpha$ of the \textit{exponential} data representation.
} \label{fig:alpha.png}
\end{figure}

Regarding the parameter $\alpha$ of the \textit{exponential} data representation, a range of candidate values has been tested. It is reminded that the default value of $\alpha$ is 24. Integer values in the range [5,90] with a step equal to 5 have been evaluated. Concerning the other settings, the Bray-Curtis distance has been used with $k=11$. The results are presented in the plot of Figure~\ref{fig:alpha.png}.
The value $\alpha=60$ provides the lowest mean validation error of 393 meters. The corresponding test set performance is characterized by a mean error of 398 meters and a median of 277 meters. All evaluated values above 60 have a difference in the achieved mean error that is less than than a meter.

\begin{figure}[!h]
\centering
\includegraphics[width=0.95\linewidth]{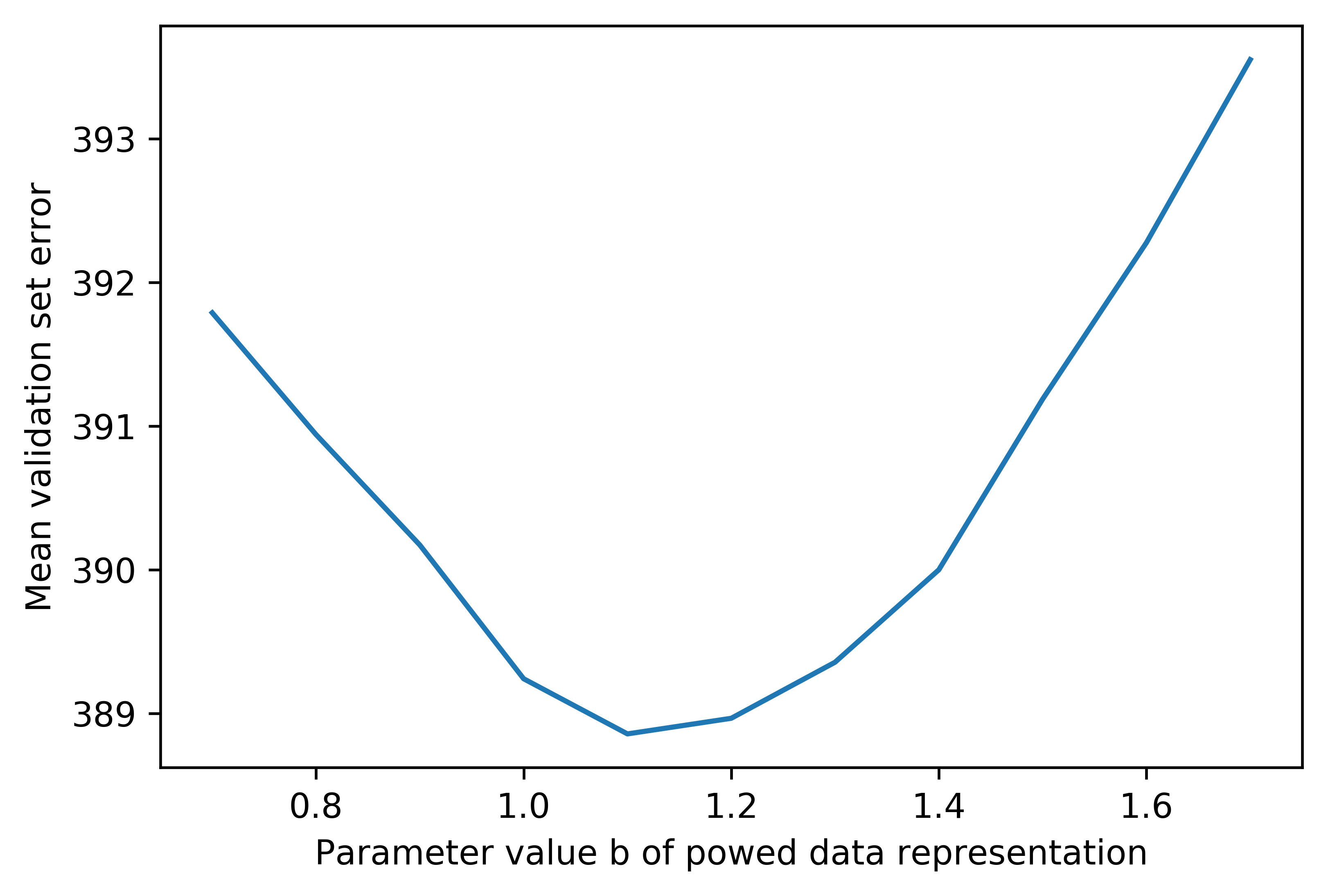}
\caption{The mean error on the validation set for different values of the parameter $\beta$ of the \textit{powed} data representation.
} \label{fig:beta.png}
\end{figure}

A similar analysis is performed for the parameter $\beta$ of the \textit{powed} data representation. The range [0.7,1.7] has been spanned with a granularity of 0.1. Concerning the other settings, the Bray-Curtis distance has been used with $k=11$. The results are presented in the plot of Figure~\ref{fig:beta.png}.
The value $\beta=1.1$ provides the lowest mean validation error of 389 meters. The mean error on the test set is 394 meters and the median 272 meters. 

\subsubsection{Choosing Distance metrics and Hyperparameter $k$} \label{sec:metrics_and_k}

\begin{table*}[h!]
\caption{Localization Error Analysis on the Validation Set, Reporting the Optimal k Value of All Distance Metrics and Four Prepossessing Strategies}\label{Table:experimental_minimum}
\centering\setlength{\extrarowheight}{2pt}
\centering
\begin{tabular}{|*{13}{c|}}
\hline
\multirowcell{2}{DistanceMetric} & \multicolumn{3}{c|}{Positive RSS} & \multicolumn{3}{c|}{Normalized RSS} & \multicolumn{3}{c|}{Exponential RSS}  & \multicolumn{3}{c|}{Powed RSS}  \\
\cline{2-13}
& \textbf{k}
& \textbf{mean}
& \textbf{median}
& \textbf{k}
& \textbf{mean}
& \textbf{median}
& \textbf{k}
& \textbf{mean}
& \textbf{median}
& \textbf{k}
& \textbf{mean}
& \textbf{median}\\ \hline
\makecell{euclidean}
& \makecell{11}  & \makecell{391} & \makecell{269} 
&  \makecell{11} & \makecell{391} & \makecell{269}
&  \makecell{11} & \makecell{391} & \makecell{270}
&  \makecell{13} & \makecell{391} & \makecell{269}\\

\makecell{manhattan}
& \makecell{12}  & \makecell{393} & \makecell{271} 
& \makecell{12}  & \makecell{393} & \makecell{271} 
& \makecell{12}  & \makecell{\textbf{392}} & \makecell{\textbf{271}}
& \makecell{13}  & \makecell{393} & \makecell{269} \\
 
\makecell{chebyshev}
& \makecell{13}  & \makecell{396} & \makecell{271} 
& \makecell{12}  & \makecell{396} & \makecell{272} 
& \makecell{10}  & \makecell{395} & \makecell{271} 
& \makecell{11}  & \makecell{395} & \makecell{270} \\ 


\makecell{hamming}
& \makecell{12}  & \makecell{499} & \makecell{385} 
& \makecell{12}  & \makecell{499} & \makecell{385} 
& \makecell{15}  & \makecell{510} & \makecell{394} 
& \makecell{13}  & \makecell{499} & \makecell{384} \\

\makecell{canberra}
& \makecell{12}  & \makecell{399} & \makecell{275} 
& \makecell{11}  & \makecell{399} & \makecell{276} 
& \makecell{11}  & \makecell{393} & \makecell{272} 
& \makecell{11}  & \makecell{399} & \makecell{276} \\
 
\makecell{braycurtis}
& \makecell{14}  & \makecell{\textbf{389}} & \makecell{\textbf{268}} 
& \makecell{13}  & \makecell{\textbf{389}} & \makecell{\textbf{267}} 
& \makecell{12}  & \makecell{\textbf{392}} & \makecell{\textbf{271}}  
& \makecell{14}  & \makecell{\underline{\textbf{388}}} & \makecell{\underline{\textbf{266}}} \\ 
\hline
\end{tabular}
\end{table*}

In this test, we examine various distance metrics and tune the hyperparameter $k$ for each distance metric. 
In these experiments, all four data representations defined in Section~\ref{sec:Dataset} were tested as a preprocessing step. The best values of the parameters $\alpha$ and $\beta$ previously found ($\alpha=60 $ and $\beta = 1.1 $) were used for the \textit{exponential} and \textit{powed} representations respectively.
The results of these tests are presented in Table~\ref{Table:experimental_minimum}. The optimal hyperparameter value $k$ and the localization error statistics reported in Table~\ref{Table:experimental_minimum} are calculated on the validation set.
In accordance to the results of other works~~\cite{Janssen_Sigfox},~\cite{IPIN2019},~\cite{TORRES_SOSPEDRA_2015}, the Bray-Curtis metric (equivalent to the S{\o}rensen metric mentioned on some of those works) is the best performing one. One the other hand, contrary to other works which use different datasets~\cite{Janssen_Sigfox},~\cite{IPIN2019},~\cite{TORRES_SOSPEDRA_2015}, the four different prepossessing steps do not seem have a significant influence on the achieved performance. This could be due to the particularities of the specific dataset. As described in Table~\ref{Table:gateways}, each fingerprint contains more than 60 out-of-range values. Since most fingerprints contain receptions by only 3 of the 68 present basestations, it seems that sharing the common set of receiving basestations is a more defining criterion for the distance between two fingerprints, than the type of prepossessing transformation performed on the RSSI values.

Apart from the metrics reported in Table~\ref{Table:experimental_minimum}, another family of distance metrics available in scikit-learn has been evaluated. This family of metrics (including metrics such as the Jaccard, Matching, Dice or Kulsinski distance) is intended to be used in boolean-valued vector spaces.
In such a setting any non-zero entry is considered as True, while zero entries are considered as False. Consequently, those representations utilize a binary type of information stating if each basestation has received the message or not. While in previous works the performance of these boolean-valued oriented methods was significantly lower compared to the metrics reported in Table~\ref{Table:experimental_minimum}, in this case their performance was not very far from the best performing one. In particular, in the current study the mean error of these boolean-valued oriented metrics lays at the range of 505-516 meters, while the best metric has a mean error of 388 meters. In a previous work~\cite{IPIN2019}, in which a different Sigfox dataset was used, the mean error of these boolean-valued oriented metrics was above 1000 meters, while the best metric gave a mean error of 317 meters. This results verify the intuitive explanation on why the differences in performance among the four preprocessing steps are so small: Due to the small percentage of receiving basestations, sharing the common set of receiving basestations is a main defining criterion for the selection of the closest fingerprints, and consequently for the resulting location estimate.

\subsection{The Extra Trees method} \label{sec:Extra_Trees}

The Extra Trees method~\cite{ExtraTrees} (standing for EXTremely RAndomized Trees), is a tree-based ensemble method, which is considered to be a variance of the Random Forests approach. Contrary to the kNN method that treats all input features as having equal importance, the creation of randomized trees utilizes a selection of the features that bring the most information.

The methods which utilize ensembles of trees have the tendency to overfit the training set to a great extent. In order to find the configuration that achieves a good performance, and which avoids overfitting the training set, a hyperparameter search on the most important hyperparameters of the method is performed.
The hyperparameter \texttt{min\_samples\_split} defines the minimum amount of samples required for splitting a node. 
Larger values of \texttt{min\_samples\_split} lead to smaller trees, higher bias and smaller variance and its optimal value hence depends in principle on the level of output noise in the dataset~\cite{ExtraTrees}.
The hyperparameter \texttt{min\_samples\_leaf} determines the minimum samples that a leaf can end up with, and has a similar smoothing effect. Lastly, the \texttt{max\_depth} determines the maximum depth of the tree. 
The selection of these hyperparameters can handle the tradeoff between an overfitting model and a model that generalizes well. Additionally, the four preprocessing strategies presented in Section~\ref{sec:Dataset} were evaluated. Lastly, the value 100 was selected for the hyperparameter \texttt{n\_estimators} of the number of trees participating in the model.

The best performance in the validation set was obtained by these values ( $\texttt{min\_samples\_split}=14$,  $\texttt{min\_samples\_leaf}=1$, $\texttt{max\_depth}= 40$), and with the use of the $powed$ data representation. The mean error on the validation set was 374 meters, and on the test set 378.
The performance of the best setting found for the Extra Trees can be seen in Table~\ref{Table:overall_performance}, along with the results of the other methods.

\subsection{Neural Nets} \label{sec:Deep}

In this subsection, the capabilities of Neural Networks in improving the accuracy of the given problem are studied. For these tests, a common feed-forward Multilayer Perceptron (MLP) architecture is studied. Since the goal is to exemplify the performance capabilities of the method, an exhaustive, systematic tuning of all related hyperparameters would be out of the scope of this study. Thus, indicative choices of the elements characterizing the architecture (number of hidden layers, neurons per layers, etc.) have been made.

A common practice that is followed in order to select the architecture and the hyperparameters of a network is to perform a two-step process. Initially, an architecture that has the capacity to overfit the training set is required. Such a network, that has the capacity to memorize to a great extend the training set, should then be able to perform well at the validation and test sets when regularization techniques are applied. Achieving a low training error in the first overfitting step lowers the bound of the achievable error of the network in the unseen data. The regularization techniques applied in the second step aim to set the model capable of achieving this generalization to unseen data.

Concerning the first overfitting step, architectures with 3, 4, 5, 6 and 7 layers have been tested. The lowest achieved error in the training set was achieved in the case of 7 layers. The mean training error achieved with 7 layers after 2000 epochs was 79 meters. Having obtained a network architecture with a considerably low training error, the effort is focused in the generalization capabilities of the model. In this direction, the effectiveness of \textit{L2 regularization},  \textit{dropout}~\cite{Dropout} and \textit{early stopping} was evaluated, with a selection of appropriate values of their relevant hyperparameters. 

The architecture selected by the first step contains 7 layers. The number of nodes in these layers were
 (1024,1024,1024,256,128,128,2). Standard methods that are commonly used as defaults in Deep Neural Network architectures have been used. More particularly, \textit{Batch normalization}~\cite{BatchNorm} was used at each layer to improve the spead and stability of the neural network. The REctified Activation Unit (\textit{ReLU})\cite{ReLU} activation function was used as the non-linearity of all hidden layers. Lastly, the commonly used \textit{Adam optimization} method~\cite{Adam} was used for the iterative update of the network weights during the training phase.

\begin{figure}[!h]
\centering
\includegraphics[width=0.95\linewidth]{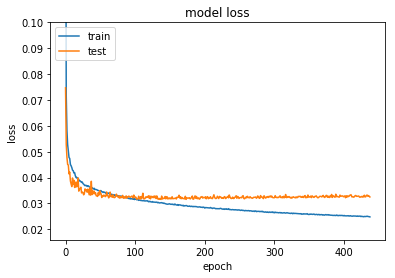}
\caption{The training and validation loss through the training epochs.
} \label{fig:zoom_loss.png}
\end{figure}

The hyperparameters that were tuned were the following. 
 \textit{Dropout} proved to be more appropriate for this setting than \textit{L2 regularization}. Thus, the setting that gave the best result is defined by a \textit{dropout rate} of $0.15$, and with $\lambda=0$ for the \textit{L2 regularization} (practically using only dropout). Moreover, \textit{early stopping} was used for selecting the stopping point of the training of the model and for avoiding overfitting. In Figure~\ref{fig:zoom_loss.png}, the training and validation losses during training are reported. While the training error goes down almost monotonically, the validation error follows a different form. It improves for a number of epochs but then it slowly but steadily increases again. \textit{Early stopping} detects this increase and after a certain number of epochs (which is a hyperparameter to be set) in which there is no improvement comparing to the lowest validation loss achieved, the training stops and the model's state at the epoch with the lowest validation loss is restored. The results obtained by the best model are reported in Table~\ref{Table:overall_performance}.

\begin{table}[h]
\caption{Localization Error Analysis of the Best Performing Configuration Found for the Three Studied Methods}\label{Table:overall_performance}
\centering\setlength{\extrarowheight}{2pt}
\centering
\begin{tabular}{|*{7}{c|}}
\hline
 \multirowcell{2}{Method} & \multicolumn{2}{c|}{Train Set} & \multicolumn{2}{c|}{Validation Set} & \multicolumn{2}{c|}{Test Set} \\
 \cline{2-7}
& \makecell{\textbf{mean}} 
& \makecell{\textbf{median}}
& \makecell{\textbf{mean}} 
& \makecell{\textbf{median}}
& \makecell{\textbf{mean}} 
& \makecell{\textbf{median}}\\ \hline

\makecell{kNN}
& \makecell{365} & \makecell{250} 
& \makecell{388} & \makecell{266} 
& \makecell{394} & \makecell{273} \\ 
\makecell{Ex.Trees}
& \makecell{283} & \makecell{202} 
& \makecell{376} & \makecell{258} 
& \makecell{380} & \makecell{261} \\ 
\makecell{MLP}
& \makecell{298} & \makecell{173} 
& \makecell{348} & \makecell{200} 
& \makecell{358} & \makecell{204} \\ 

\hline

\end{tabular}
\end{table}

\section{Conclusions} \label{sec:Conclusions}

In this work, a comparison of three different types of fingerprinting methods is performed, in the context of an outdoor LoRaWAN setting. A public dataset has been used and the performance of the tested methods has been characterized. The results obtained by these methods are reported in Table~\ref{Table:overall_performance}. The selection and evaluation of the models has been carried out with the common train/validation/test set split. These three subsets of the original dataset are made publicly available, in order to facilitate the reproducibility and the comparability of the results.

The best performing model proposed in this work achieves a mean error of 348 meters and median error of 200 meters, on the validation set. The corresponding performance in the test set is 358 meters of mean error, and 204 meters of median error. To the best of our knowledge, the only other work which presents an evaluation of the position accuracy of positioning methods using this dataset, is the work the presented this dataset~\cite{Sigfox_Dataset}. In that work, the performance of the kNN method was tested, in which the Euclidean distance was used and the hyperparameter $k$ was tuned, with the goal to exemplify the usage of the dataset. 

A comparison with the performance of previous works would not be strictly consistent, even if the same dataset has been used, since the validation and test sets would not be the same. Nevertheless, in the original study~\cite{Sigfox_Dataset} similar results are observed comparing with the ones given by the kNN tests of the current study. The mean and median error achieved by Aernouts et al.~\cite{Sigfox_Dataset} is 398 and 273, while in Table~\ref{Table:overall_performance} it can be seen that kNN achieves 394 and 273 meters respectively. Thus, despite the fact that different test sets have been used, the performance differences are quite small.

The best performing method of this study was the neural network approach, where a Multilayer Perceptron (MLP) was used. This method achieved a 358-meters mean error, and 204 meters median error. These results constitute a 41 meters (and a 10\%) improvement comparing to the mean error presented in the previous work~\cite{Sigfox_Dataset}. Similarly, the median error shows a 69-meters (and a 25\%) improvement. The difference in the improvement percentage between the mean and the median error (10\% and 25\% respectively) can be explained by the presence of outliers, which affect the mean but not the median error. Thought the accuracy of most position estimates significantly improves, the presence of outliers with high error keeps the percentage improvement of the mean error lower.

\section*{Acknowledgements}
\label{sec:Acknowledgements} 
This work was funded by the Commission for Technology and Innovation CTI, of the Swiss federal government, in the frame of the OrbiLoc project (CTI 27908.1 PFES-ES).

\balance

\bibliographystyle{IEEEtran}
\bibliography{bibliography}

\end{document}